# From Binary to Bilingual: How the National Weather Service is Using Artificial Intelligence to Develop a Comprehensive Translation Program

Joseph E. Trujillo-Falcón,[a,b] Monica L. Bozeman,[c] Liam E. Llewellyn,[d] Samuel T. Halvorson,[a,e] Meryl Mizell,[f] Stuti Deshpande,[g] Bob Manning,[h] Chris Rohrbach,[i] Ian Blaylock,[j] Angel Montanez,[k] and Todd Fagin[l]

[a] *Department of Climate, Meteorology & Atmospheric Sciences, University of Illinois Urbana-Champaign, Urbana, IL*

[b] *Department of Communication, University of Illinois Urbana-Champaign, Urbana, IL*

[c] *NOAA/National Weather Service Office of Central Processing, Silver Spring, MD*

[d] *Department of Geography & Geographic Information Science, University of Illinois Urbana-Champaign, Urbana, IL*

[e] *Department of Atmospheric Sciences, University of North Dakota, Grand Forks, ND*

[f] *Pace University, Pleasantville, NY*

[g] *NOAA/National Weather Service Office of Observations, Silver Spring, MD*

[h] *LILT, Emeryville, CA*

[i] *NOAA/National Weather Service, Columbia, SC*

[j] *NOAA/NWS Southeast River Forecast Center, Peachtree City, GA*

[k] *Cooperative Institute for Research in the Atmosphere, Colorado State University, Fort Collins, CO*

[l] *Center for Spatial Analysis, University of Oklahoma, Norman, OK*

*Corresponding author*: Monica Bozeman, monica.bozeman@noaa.gov

ABSTRACT

To advance a *Weather-Ready Nation*, the National Weather Service (NWS) is developing a systematic translation program to better serve the 68.8 million people in the U.S. who do not speak English at home. This article outlines the foundation of an automated translation tool for NWS products, powered by artificial intelligence. The NWS has partnered with LILT, whose patented training process enables large language models (LLMs) to adapt neural machine translation (NMT) tools for weather terminology and messaging. Designed for scalability across Weather Forecast Offices (WFOs) and National Centers, the system is currently being developed in Spanish, Simplified Chinese, Vietnamese, and other widely spoken non-English languages. Rooted in best practices for multilingual risk communication, the system provides accurate, timely, and culturally relevant translations, significantly reducing manual translation time and easing operational workloads across the NWS. To guide the distribution of these products, GIS mapping was used to identify language needs across different NWS regions, helping prioritize resources for the communities that need them most. We also integrated ethical AI practices throughout the program's design, ensuring that transparency, fairness, and human oversight guide how automated translations are created, evaluated, and shared with the public. This work has culminated into a website featuring experimental multilingual NWS products, including translated warnings, 7-day forecasts, and educational campaigns, bringing the country one step closer to a national warning system that reaches *all* Americans.

SIGNIFICANCE STATEMENT

The National Weather Service has recently trained artificial intelligence to translate weather products into multiple languages. In this article, we review the development of the software, introduce how we plan to operationalize the system, outline how we ensure ethical practices, and provide a first look at NWS products being developed by the agency.

## 1. Introduction

The United States is best described as a kaleidoscope of cultures, a country where more than 350 languages coexist. As of 2022, nearly 68.8 million Americans—about 1 in 5—speak a language other than English at home (U.S. Census Bureau 2024). Yet, English remains the primary language used in weather warning communication during hazardous weather disasters (Trujillo-Falcón et al. 2021). This raises an important question: if our country speaks multiple languages, why does our weather communication system rely on only *one*?

Limited access in one's primary language can have serious, even life-threatening, consequences. The earliest documented casualty related to language inequities in a government report dates back to 1970, when a tornado struck Lubbock, Texas, disproportionately impacting their Spanish-speaking community (U.S. Department of Commerce 1970). Since then, research has consistently linked language barriers to increased vulnerability across a range of extreme weather disasters, including hurricanes (Benavides 2013), floods (Maldonado et al. 2016), tornadoes (Gaviria Pabón et al. 2025), and wildfires (Méndez et al. 2020). Individuals with Limited English Proficiency (LEP)[1] are particularly vulnerable, as they frequently lack access to lifesaving information in a language they can understand (Aguirre 1988; Trujillo-Falcón et al. 2024a). Research has established that without a systematic approach to delivering *dialect-neutral translations and culturally resonant communication*, these inequities will persist and continue to impact the most vulnerable groups across the country (Trujillo-Falcón et al. 2021, 2022, 2024b).

Addressing the aforementioned inequities requires institutional and human capacity to support multilingual communication at scale. Recent workforce trends suggest that this capacity is beginning to emerge within the weather, water, and climate enterprise (Medina Luna et al. 2019), following a similar transition that occurred previously in the medical field (Ortega and Prada 2020). In recent years, operational and research communities across geosciences have seen a growing number of professionals from diverse backgrounds, including individuals with multilingual and multicultural expertise (Morales et al. 2023). This shift creates new opportunities to strengthen bilingual communication for communities that speak various non-English languages across the country. At the same time, recent studies show that bilingual professionals across the field lack access to foundational dictionaries, standardized terminology, and reference materials needed to consistently translate technical meteorological information across languages (MacPherson-Krutsky et al. 2025; Trujillo-Falcón et al. 2025). Addressing these needs is essential for supporting bilingual meteorologists and ensuring that developing translation technologies are responsibly and consistently integrated into operational workflows.

[1] Limited English Proficiency (LEP) refers to individuals in the United States who do not speak English as their primary language and have a limited ability to read, speak, write, or understand English (U.S. Census Bureau n.d.). The U.S. Census identifies individuals as LEP when they report speaking English less than "very well."

At the institutional level, the NWS has begun addressing the critical need for multilingual communication strategies. These efforts originated in San Juan, Puerto Rico, where bilingual communication has been a core part of operations since the mid-1960s (Negrón-Hernández and López 2025). Since 2014, operational forecasters across the NWS have formed volunteer translation teams to provide Spanish-language communication across the agency (Trujillo-Falcón et al. 2021). Over time, other languages, such as Samoan and Vietnamese, have also been incorporated to meet the needs of some local Weather Forecast Offices (WFOs) (Lefale 2010). Despite these initiatives, an internal review found that the NWS still lacks a comprehensive strategy to effectively communicate with multilingual groups, including those with LEP or communication disabilities (NWS 2023).

In response to these challenges, the NWS launched the AI Language Translation Program in collaboration with LILT, a company specializing in AI-powered translation services (NOAA 2023). The goal of the partnership is to develop translation models built specifically for weather communication, capable of automatically translating technical jargon, scientific terminology, and forecast information with a high degree of accuracy and nuance beyond what is typically achievable with open-source translation tools. These purpose-built translation models go beyond word-for-word translation by emphasizing message clarity, plain language, and culturally appropriate communication that is widely understood. Initially focused on Spanish, the NWS AI Translation Program has since expanded to include languages such as Vietnamese, Mandarin, Samoan, and French, representing some of the largest LEP communities in the United States (Bozeman et al. 2024a). The initiative has enhanced translation accuracy to reduce the risk of life and property loss, ease the operational workload on forecasters, speed up translation times, and create a scalable and systematic approach to multilingual communication across the agency. These efforts have led to the development of an experimental website, where users can preview the future of NWS multilingual operations (weather.gov/translate).

This article provides an overview of the NWS AI Language Translation Program. Section 2 details the development of the neural translation software, its integration into NWS systems, and the methodology used to ensure accurate translations across multiple languages, including an example of a resulting experimental product. Section 3 covers the product's implementation across multiple WFOs, using insights from GIS and U.S. Census Bureau data. We highlight the development of a GIS dashboard designed to identify areas where the neural translation software could offer the greatest benefit. Section 4 addresses ethical AI and the steps our team took to ensure ethical practices throughout the development of the translation tool. Finally, Section 5 offers a first glimpse into the future of a multilingual NWS, showcasing an experimental website and plans to incorporate social and behavioral science research moving forward.

## 2. The Development of the NWS AI Language Translation Program

### *a. Background and Motivation*

Before artificial intelligence tools emerged, bilingual operational forecasters led grassroots translation efforts within the NWS. In 2014, the agency advanced these efforts by creating volunteer translation teams across its offices (Trujillo-Falcón et al. 2021). These teams advanced access to weather information for non-English-speaking communities across the country. However, the process depended heavily on the availability of bilingual personnel and often overwhelmed forecasters, as translation became an added responsibility on top of their core operational duties. The absence of a standardized translation protocol further complicated these efforts, leading to inconsistencies across NWS offices. In particular, dialectical variations in Spanish posed persistent challenges, as translations frequently prioritized literal word-for-word conversions over conveying accurate meaning (Bitterman et al. 2023; Trujillo-Falcón et al. 2022). Collectively, these limitations highlighted the need for a more systematic, scalable approach to multilingual communication across the NWS.

To accomplish the needs of the NWS to warn and communicate with LEP communities, the agency established the NWS AI Language Translation Program in early 2021. The program set out with three core objectives: (1) support bilingual staff by developing assistive translation tools to streamline and speed up the translation process; (2) transition local, office-level translation efforts to a unified, enterprise-wide system; and (3) fully automate translation capabilities to serve offices lacking bilingual personnel (Bozeman et al. 2024a). To begin this process, the NWS Office of Central Processing began testing AutoML, a commercial AI translation model developed by Google, to better understand the capabilities of artificial intelligence within the translation industry (Bozeman et al. 2024b). Through this pilot, the team identified critical needs, including compiling and formatting a parallel corpus of previously translated NWS content to train a Spanish-language AI model, and resolving challenges related to the dissemination of translated text. These early efforts made it clear that the agency could not build and scale a comprehensive AI translation system without external expertise (Bozeman et al. 2024b).

To address this challenge, the NWS issued a Request for Information (RFI) in December 2021 to assess available translation and AI solutions in the private sector (NOAA 2021). Based on the market research and RFI responses, the agency selected LILT as a potential provider (Bozeman et al. 2024b). LILT's platform offers adaptive, real-time training of AI language translation models, supports human-in-the-loop workflows, and can source professional linguists and terminologists[2] for its clients (Stiefel and Manning 2024). Additionally, LILT's customer base and prior work with the U.S. Air Force through the Small Business Innovation Research program demonstrated its ability to handle sensitive information while maintaining a high level of security (Bozeman et al. 2024a). Before formalizing the partnership, the NWS launched a nine-month pilot project to evaluate LILT's suite of services in an operational context.

*b. Developing the NWS/LILT Neural Translation Process*

To put LILT's technology into practice and establish the foundation for an enterprise-wide translation solution, a 9-month pilot focused on supporting manual Spanish translations conducted by the San Juan, Puerto Rico WFO (hereafter referred to as "SJU"), a team with over three decades of experience of translating weather products (see Bozeman et al. 2024a for detailed documentation of the bilingual forecasters involved in the early pilot work). The pilot centered on three textual NHC products frequently used during hurricane season: (1) the Tropical Public Advisory (TCP), (2) the Tropical Forecast Discussion (TCD) containing the NHC "Key Messages," and (3) the Tropical Weather Outlook (TWO). Using previously translated English–Spanish product pairs from the 2017–2020 hurricane seasons, the NWS trained the AI model to handle their domain-specific language. This phase of the project explored how AI translation could integrate into current NWS systems, assess whether it could reduce the translation workload for forecasters, and determine if LILT's adaptive model could maintain long-term accuracy and reliability.

---

[2] In this paper, we distinguish between *linguists* and *terminologists* based on their roles in the translation workflow. Linguists focus on language broadly, including structure, meaning, and usage, and often support translation through applied linguistic analysis. Terminologists, by contrast, utilize their additional specialized training to identify, define, and standardize domain-specific vocabulary, ensuring consistency and accuracy of technical terms across documents, languages, and dialects. In this project, linguists contributed to language quality and translation processes, while terminologists focused on managing and standardizing specialized weather terminology used to train and evaluate the translation models.

During the SJU pilot, developers scoped and optimized the translation code to align with NWS operational infrastructure for a human-in-the-loop translation workflow. The NWS translation processor reformatted raw NWS English text files to make them compatible with the translation engine, then uploaded them through the LILT API. Within the LILT online Computer-Assisted Translation (CAT) tool, which uses LILT's Adaptive Neural Machine Translation (NMT) system (hereafter referred to as the "LILT Platform" or "LILT interface"), an SJU forecaster interacted with the model to translate each file. The translated product returned to NWS infrastructure, where forecasters were provided a one-hour window to review and edit the translation before the system automatically reformatted and disseminated the product. SJU's bilingual forecasters dynamically trained the AI model in near real time as each NHC product was issued and translated within the LILT Platform.

To train the AI, the LILT model first generated a draft translation, which the linguist edited. Each time the linguist adjusted an AI-suggested translation, the model immediately reevaluated the rest of the sentence based on those edits. After completing the document translation, secondary reviewers refined the text and marked each sentence as "reviewed," finalizing the translation and updating the model to improve future outputs. These human reviewed sentences were then included in the model memory bank. During future translation tasks, LILT applied these 100% memory matches—sentences that exactly matched previously human-reviewed translations—as verified reference translations, rather than generating new AI suggestions. This human-in-the-loop AI workflow drew on the technical expertise of NWS staff, strengthening accuracy and trust in the resulting automated translations.

As bilingual forecasters continued to work and translate within the LILT platform, the number of human-verified reviewed segments in the model's memory grew quite large over the 9-month pilot period. The system automatically inserted these verified memory matches into new product translations, enabling forecasters to focus only on unmatched sentences. As a result, translation time dropped substantially during the 2022 hurricane season and improved even further in 2023 (Table 1). According to LILT's scoring metrics, the Spanish model achieved a Word Prediction Accuracy (WPA) score of 95% for the three products translated by SJU, reflecting alignment between model outputs and human-in-the-loop edits (Table 1).

| NWS Text Product | Pre-LILT Translation Time | 2022 LILT Unadapted Spanish model WPA Score | 2022 Translation Time | 2022 Adapted WPA Score | 2023 Translation Time | 2023 Adapted WPA Score |
|---|---|---|---|---|---|---|
| Tropical Weather Outlook (TWO) | ~10–30 mins | 65% (estimated) | 2–5 mins | 95% | 2–3 mins | 99% |
| Public Tropical Cyclone Product (TCP) | ~45 mins–1 hour | 65% (estimated) | 10–15 mins | 95% | 5–7 mins | 99% |

**Table 1**. SJU bilingual forecaster's time to translate and WPA score utilizing the LILT AI translation system.

The NWS demonstrated through the SJU pilot that training AI models using the LILT platform was more effective and scalable than post-editing machine translations and re-ingesting corrected outputs into the Google AutoML model. Encouraged by these results, the NWS advanced toward its objectives and launched additional pilots to expand product coverage, standardize terminology, and further automate and streamline the translation process.

*c. Developing a General Framework for Translation in Multiple Non-English Languages*

Building on the SJU pilot, the NWS began scaling its automated system with the goal of producing translations that remain clear and consistent across regional language variations, as recommended by Trujillo-Falcón et al. (2021). The program pivoted to creating a new "National Spanish" model based on the successes of the SJU pilot, designed to produce Spanish that is broadly understood across Spanish-speaking communities and dialects in the United States. In addition to the specialized terms that SJU forecasters entered into the LILT system during the hurricane season pilot, LILT's professional terminologists conducted an in-depth analysis of the AI model's translation memory, built from SJU human-verified translations during the 2022 hurricane season. These terminologists cross-referenced the extracted terms with several authoritative Spanish-language and multilingual meteorological resources, including the following:

- *Diccionario de americanismos* by the Asociación de Academias de la Lengua Española (n.d.)
- *Diccionario de la lengua española* by the Real Academia Española (n.d.)
- TERIMUM Plus® terminology database by the Government of Canada (n.d.)
- METEOTERM/UNTERM glossaries by the World Meteorological Organization and the United Nations (n.d.)

Terminologists combined this corpus with the existing English-to-Spanish Glossary from the NWS (n.d.) to perform a word and phrase frequency analysis using SketchEngine, a linguistic tool widely used by their field. SketchEngine (n.d.) allows for the examination of large corpora to identify typical usage patterns, rare constructions, and emerging terminology in a given language. From this terminology analysis process called "term harvesting," the terminologists developed an overall list of specialized terms used within NWS content as a basis for the Spanish and future AI model's term overrides known as a "termbase," or colloquially known as a "multilingual glossary."

During the term harvesting process, the terminologists also advised the NWS on which terms from the list had multiple meanings and/or valid dialectal variants, and from which countries those terms originated. Reference material from the *Diccionario de americanismos* greatly contributed to this effort by identifying which Spanish-speaking countries–and therefore dialects–used alternate terms that should be considered when creating NWS-translated content appropriate for audiences in the U.S. and the Caribbean (Real Academia Española n.d.). For example, the term historically used by the San Juan forecasters for "thunderstorm" was "tronada." However, through the extensive term research performed by LILT and by polling multiple NWS bilingual staff with diverse Spanish heritages, we found that Cubans and Mexicans, for instance, did not recognize this regional term but understood the more generalized "tormenta" or "tormenta eléctrica". As a result, the NWS Spanish AI model adopted "tormenta eléctrica" or "tormenta," where appropriate, into its termbase for use by bilingual staff across the agency. This cross-cultural terminology research played a critical role in refining the corpus and ensuring culturally consistent translations across dialects.

In addition to the term harvesting process, the NWS combined multiple techniques to adapt LILT LLMs on NWS content and terminology (Fig. 1). For instance, LILT's professional computational linguists, also trained in data science, conducted advanced quality control (QC) on the historical SJU corpus to train the new Spanish translation model. This QC process identified and removed over 9,000 instances of unintended model behavior, such as inconsistent phrasing or outdated terminology across approximately 44,000 sentences (Fig. 1). Many of these inconsistencies stemmed from minor human errors or changes in San Juan's translation preferences over time. To further strengthen the model, the LILT team generated approximately 200 sentences of synthetic data to supplement the training set (Fig. 1). These synthetic sentences were designed to mimic common forecast language which included variable elements such as days of the week, temperature ranges, rainfall amounts, wind speeds, and sky conditions. For example, sample sentences included: "[*Today* | *Tomorrow*] will be [*sunny* | *cloudy*] with a high temperature in the low [*60*]s." or "[THURSDAY] NIGHT…[Partly cloudy]. Lows in the upper [40]s. [North] winds around [5] mph." These types of sentences are frequently found in NWS Zone Forecast Products issued by WFOs and in TWOs produced by the NHC (e.g., "Formation chance through 5 days…[low]...[20] percent, "as seen in Fig. 2).

**3 Layers of Language Model Adaptation**

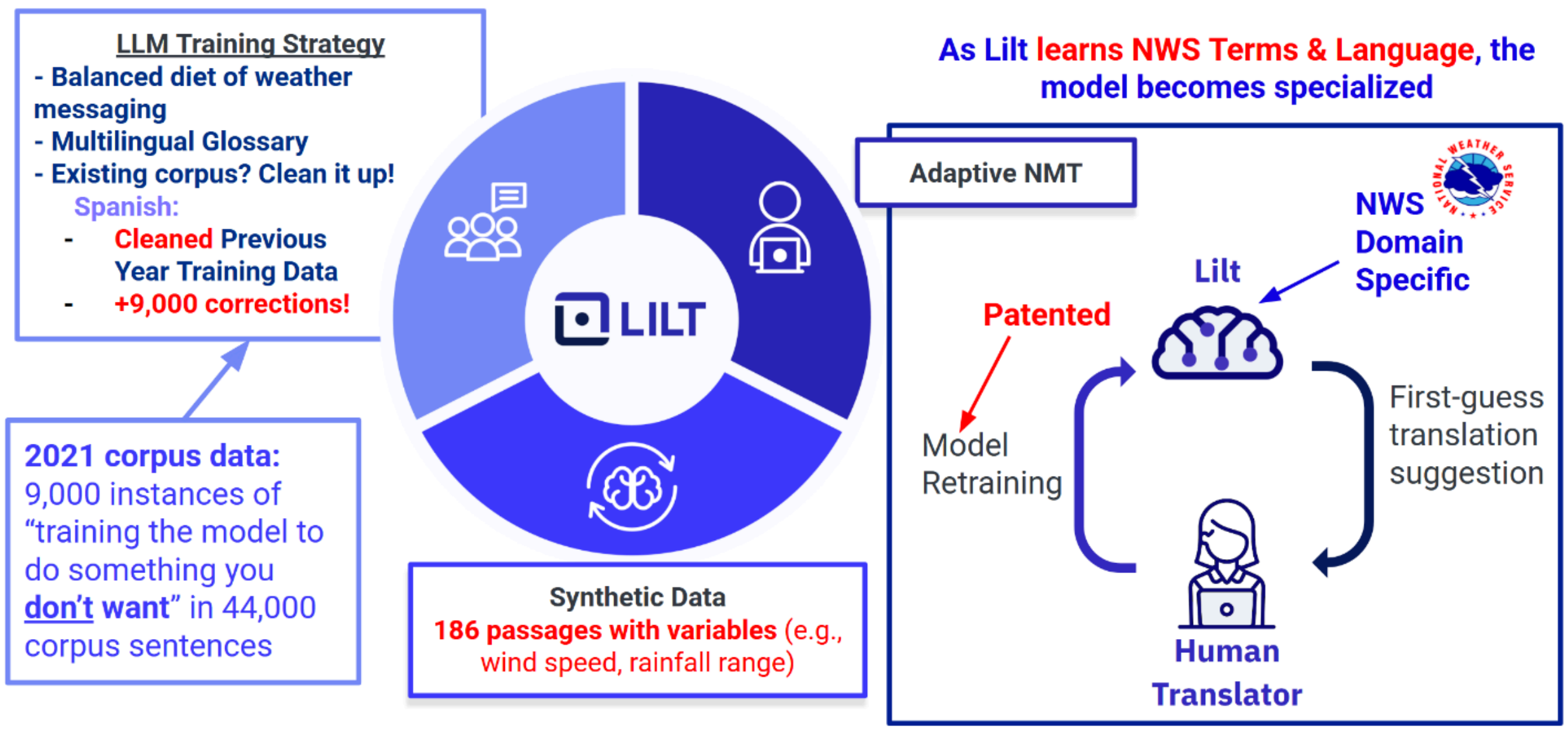

**Fig. 1.** Visual overview of the multiple layers of AI model initial training and adaptation work completed by LILT computational linguists, terminologists, NWS AI Language Translation Team members and NWS bilingual forecasters serving as human-in-the-loop translators operating on the LILT platform.

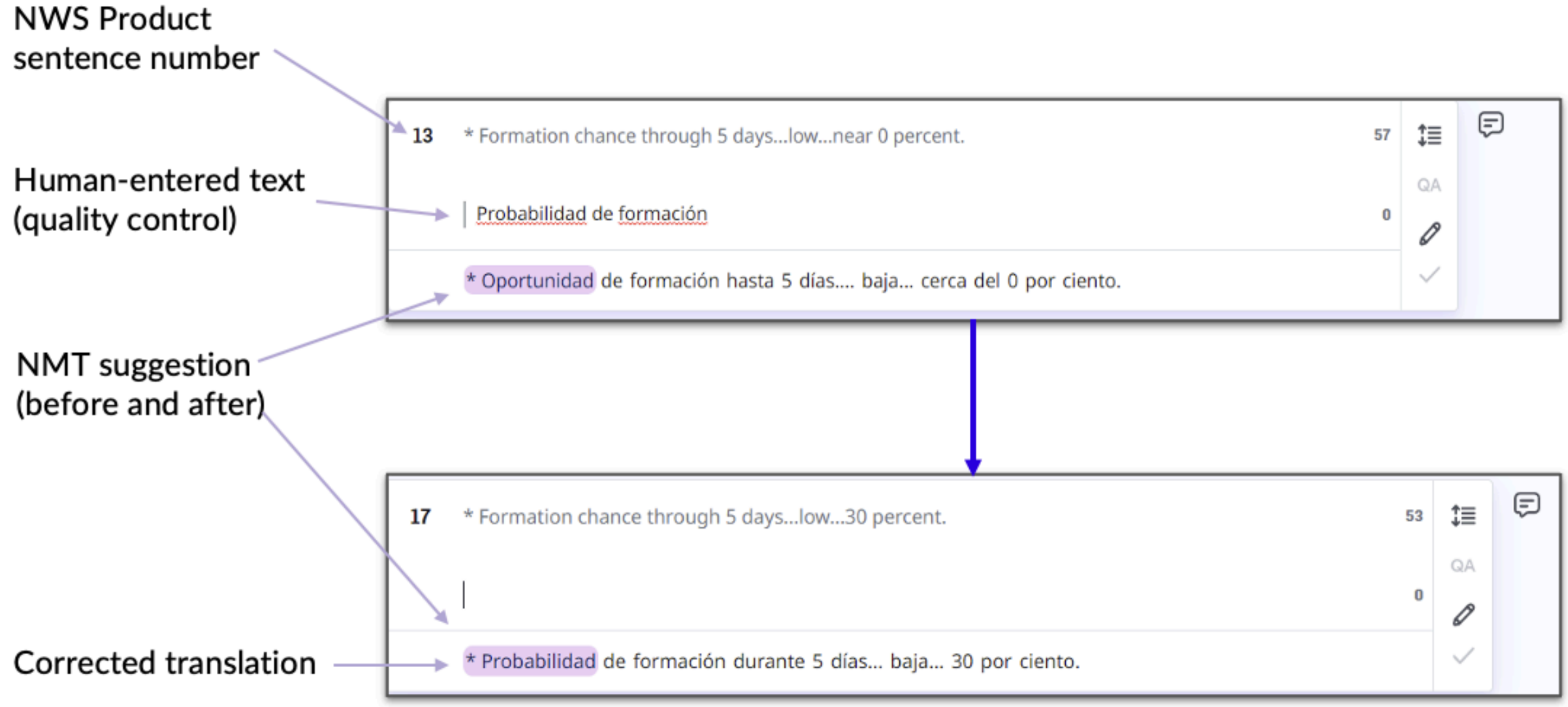

**Fig. 2.** Real-time LLM training within the LILT interface for an NHC TWO.

Building on lessons learned from early and subsequent pilot efforts (NOAA n.d.), our current framework integrates a human-in-the-loop approach to fine-tune a large language model and encompasses multiple non-English languages, such as Spanish, Chinese, Vietnamese, French, and Samoan (Fig. 3). Within this framework, the custom NWS LLM builds upon the LILT base model by initially training on a corpus of roughly 60,000–80,000 words of NWS domain-specific content to establish a strong baseline. After model deployment, professional translators and bilingual meteorologists continue to provide quality control and minor linguistic adjustments as well as completing strategic AI-assisted translation jobs throughout specific months of the year. These carefully selected translation jobs include NWS content from different seasons, hazard types, and regions, capturing both significant and routine weather events along with the language used to communicate them nationwide. We colloquially refer to these jobs as the “balanced diet,” an intentionally selected mix of monthly translation tasks used to ensure that models do not go stale. Each human-verified translation feeds back into the system as supervised training data, strengthening the model’s ability to reflect the terminology, tone, and structural preferences of the issuing agency (see purple box in Fig. 3). To meet the urgency of real-time warnings, translation requests can also be routed directly to the LLM for autonomous processing (see green box in Fig. 3), where it draws on stored human-curated translation memory matches and, when needed, generates new outputs using neural machine translation techniques.

# Translation Tools & Path to Public

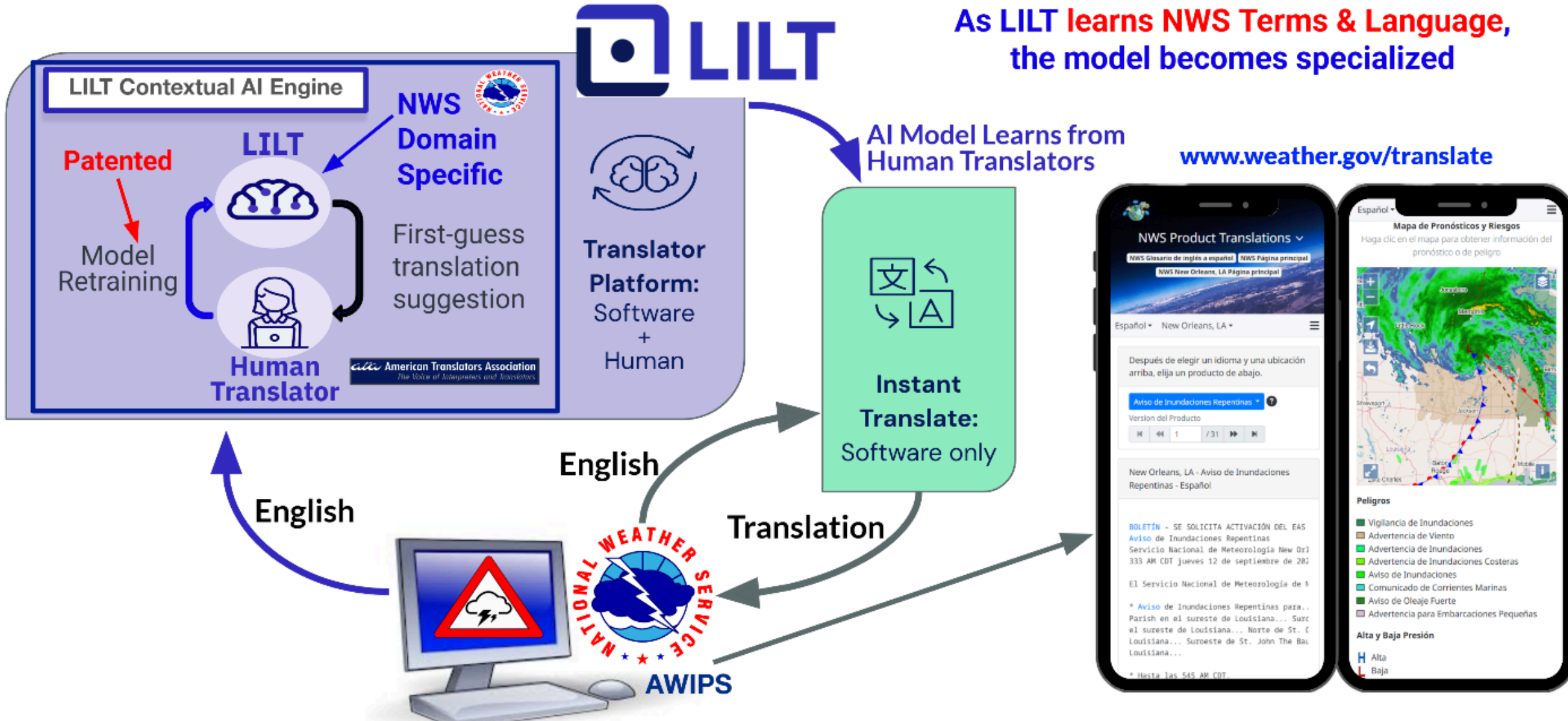

**Fig 3.** Workflow for NWS text products translated by the LILT API and posting to the public NWS experimental website weather.gov/translate.

*d. Verification Methods*

Verification methods are essential for ensuring the accuracy and reliability of translations. They also help confirm that the intended meaning of the original message is preserved, an especially critical concern in weather communication (Trujillo-Falcón et al. 2022). In linguistics, mathematical and computational metrics have long provided a foundation for evaluating translation accuracy. One of the earliest and most widely used is the Levenshtein Distance, or 'edit distance,' which measures similarity between two strings of text by calculating the number of insertions, deletions, or substitutions required to transform one into the other (Levenshtein 1966). This metric quantifies the difference between a machine translation and a reference text, laying the groundwork for newer metrics that capture other dimensions of translation quality.

Translation vendors often implement internal techniques to evaluate the quality of output generated by their LLMs. Unfortunately, in an effort to differentiate themselves from other vendors, these evaluation methods are frequently proprietary, posing a challenge to compare translation accuracy across platforms like Google Translate, Gemini, ChatGPT, LILT, and others. While some platforms support the use of external vendor models and apply a consistent scoring framework across them, this functionality is not yet universally available.

During the NWS's vendor selection process for AI translation, we found that most companies offering model evaluation metrics, when available to customers, use proprietary adaptations of the traditional "edit distance" concept. For instance, LILT developed the WPA score to measure how closely an LLM's word predictions align with human-in-the-loop edits during translation to support supervised AI training (as seen in Table 1). The WPA score is central to how LILT tracks translation quality during model refinement, but it is not calculated for outputs from LILT's Instant Translate tool (see green box in Fig. 3), which generates text without human oversight. Because the NWS relies on Instant Translate for automated translations, this lack of scoring capability presented the agency with three challenges: (1) how to evaluate the quality of translations from Instant Translate, (2) how to produce a detailed "report card" identifying which message types (e.g., tornado vs. hurricane warnings) performed well or needed improvement, and (3) how to ensure that scientific nuance and meaning were preserved across languages.

The NWS therefore consulted with computational linguists to develop a custom approach and determine the best linguistic scoring algorithms applicable to our requirements. This approach involves devising a way to score both the overall performance of the entire model and the translations on a sentence-by-sentence basis to meet NWS needs for such detailed reporting. Scoring at the sentence level provides confidence in the quality of the translation and is further useful in deducing which file or NWS product types require further investigation and highlight where additional model training or human linguistic adjustments may be necessary. Following expert guidance from computational linguists, we selected five linguistic scoring algorithms for evaluation and comparison (Table 2).

| **Metric** | **Core Equation** | **High-Level Explanation** |
|---|---|---|
| Bilingual Evaluation Understudy Score (BLEU) | $p_n = \frac{\sum_{C \in \{Candidates\}} \sum_{n-gram \in C} Count_{clip}(n-gram)}{\sum_{C' \in \{Candidates\}} \sum_{n-gram' \in C'} Count_{clip}(n-gram')}$ | Scores how similarly machine translations match human translations using a word-by-word n-gram overlap comparison (Papineni et al. 2002) |
| Character F-score (ChrF++) | $chrF\beta = (1 + \beta^2) \frac{chrP * chrR}{\beta^2 * chrP + chrR}$ | Evaluates translation quality at the character level by comparing character |
| | | overlaps (Popović 2015) |
| Cross-lingual Optimized Metric for Evaluation of Translation (COMET) | $e_{x_j} = \mu E_{x_j}^{T} \alpha$ | A neural framework for machine translation evaluation based on contextual accuracy and fluency (Rei et al. 2020) |
| Translation Edit Rate (TER) | $TER = \frac{\# \, of \, edits}{average \, \# \, of \, reference \, words}$ | Measures the number of edits needed to make a target translated sentence look the same as the original reference sentence (Snover et al. 2006) |
| Fuzzy String Matching (Fuzz) | $C^{s}_{\|\|x\|\|-s+1} << P_{s}(x) << C^{s}_{\|\|x\|\|+s+1}$, (1) <br> $\sum_{i=0}^{s} C^{i}_{n} 2^{2-i} \ll Q_{s}(x) \ll \sum_{i=0}^{s} C^{i}_{n} 2^{s-i}$ (2) | Evaluates the similarity between strings using edit distance (Levenshtein 1966) |

**Table 2.** Linguistic scoring algorithms evaluated and high-level summary

Initially, the BLEU score was implemented as the primary measure of translation quality. BLEU is widely used across the translation industry to compare "apples to apples" across vendor LLMs and measures the degree of n-gram[3] overlap between machine-generated and human reference translations (Papineni et al. 2002). However, it soon became clear that BLEU alone was insufficient, especially for capturing fluency and semantic accuracy. For example, BLEU applies a brevity penalty that may unfairly penalize shorter, but semantically correct translations. To mitigate this, the NWS applied smoothing techniques, which improved performance to some extent. Despite these approaches, additional metrics were needed to more effectively capture meaning and contextual nuance (Vashee 2019).

Recognizing the need for deeper evaluation, additional consultation with computational linguists resulted in more metrics added to the workflow, including ChrF++, COMET, TER, and Fuzz score (Table 2). ChrF++ and COMET offered improved sensitivity to semantic meaning

[3] An n-gram is defined as an adjacent sequence of a certain number of words. For example, in the sentence "The weather is dangerous," the following n-grams can be extracted:

- Unigrams (1-word sequences): "The," "weather," "is," "dangerous"
- Bigrams (2-word sequences): "The weather," "weather is," "is dangerous"
- Trigrams (3-word sequences): "The weather is," "weather is dangerous"

and character-level nuance. COMET evaluates the semantic similarity between a machine-translated output and the source text through pre-trained neural models. It has consistently outperformed traditional automatic metrics in predicting human judgment of translation quality (Rei et al. 2020). TER calculates the number of edits needed to match a reference, while the Fuzz score uses fuzzy string-matching algorithms like Levenshtein Distance to assess similarity. Although both TER and Fuzz measure edit distance, we observed inconsistencies between their outputs due to different underlying formulas. Moreover, we found that metrics such as ChrF++, TER, and Fuzz often produced artificially low scores when encountering non-translatable elements like proper names and numbers. Performance also varied across target languages, with some metrics performing better in specific language pairs than others. These insights reinforced the decision to rely primarily on BLEU and COMET scores, while keeping the other metrics available for supplemental analysis.

In addition to these metrics, the NWS adopted a complementary strategy common in the translation industry known as back translation (or reverse translation). In this method, a translated message is retranslated back into English, using a separate, untrained model, and compared to the original source text. This allows both bilingual and non-bilingual reviewers to identify discrepancies and investigate potential errors, especially when dealing with technical language (Argo Translation 2024). An important note about back translations is that they typically do not sound as smooth or fluent as an initial translation. Back translations are strategically used as a tool to assess meaning via literal unbiased reverse translations revealing the "worst-case scenario" where the back translation is not meant to be massaged for fluency. This methodology is particularly useful for preserving scientific accuracy in meteorological terms, ensuring that translations go beyond literal word-for-word renderings. By using back translations, the NWS can make progress towards meeting goal three to ensure that the meaning of scientific concepts is retained and not merely providing literal translations that could be confusing to an LEP audience.

With the metrics and methodology in place, the NWS then developed an automated dashboard to track and display translation performance. This prototype captures real-time metadata from LILT translation jobs, including language pair, job ID, and processing time. Using REST API calls, the dashboard retrieves segment-level data and stores the original and translated sentence pairs in a cloud-based database (here, the NWS uses an AWS DynamoDB). It then requests back translations and calculates all translation metric scores by comparing the original English source with the reverse-translated output, representing a “worst-case scenario” benchmark. Fig. 4 depicts the process workflow of the back translation, scoring and dashboard visualization process, while Tables 3 and 4 illustrate an example of evaluating a translation of the English phrase “Tropical Weather Outlook” translated to Spanish and French. This example is intended to illustrate how different evaluation metrics respond to back translation relative to a preferred human translation of a key term, rather than to assess overall model performance. Because the example phrase in Table 3 is intentionally short, n-gram–based metrics such as BLEU yield artificially low scores in Table 4, a known limitation of these metrics when applied to short texts.

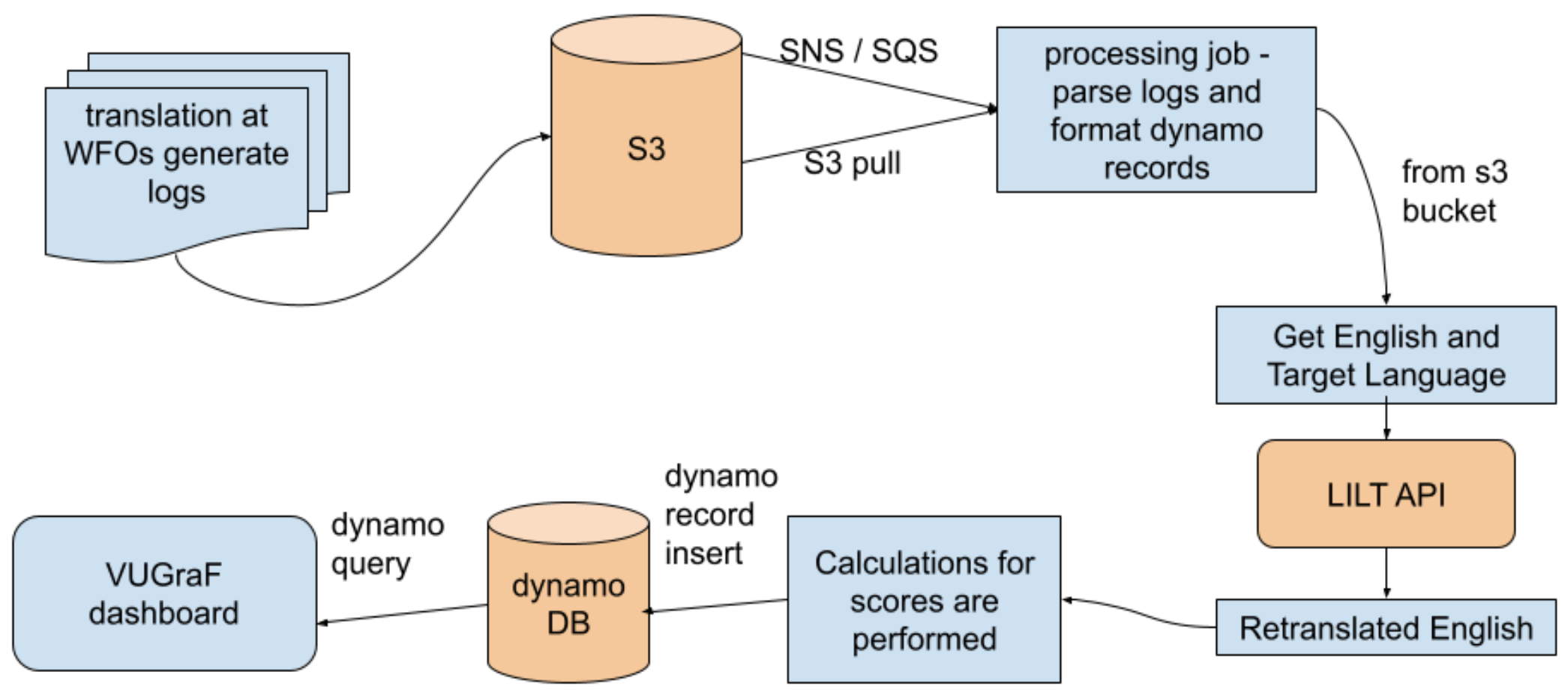

**Fig. 4**. Automated workflow for the NWS back translation process and scoring metric calculations for display and analysis within the accuracy dashboard.

| **Original English Phrase: “Tropical Weather Outlook”** | | |
|---|---|---|
| | **Spanish** | **French** |
| **Human translation (for reference)** | “Perspectiva sobre las Condiciones del Tiempo en el Trópico” | “Aperçu des conditions météorologiques tropicales” |
| **Machine translation** | “Perspectiva sobre las Condiciones del Tiempo en el Trópico” | “Prévisions météorologiques tropicales” |
| **Reverse Translation** | “Perspective on Weather Conditions in the Tropics” | “Tropical weather forecast” |

**Table 3**. Example of how the phrase "Tropical Weather Outlook" is translated through human vs machine translation and then compared to its English back translation processed through LILT untrained reverse directional models.

| | **BLEU** | **Fuzz** | **chrF++** | **COMET** | **TER** |
|---|---|---|---|---|---|
| **Spanish Scores** | 6.567/100 | 42/100 | 29.626/100 | 0.723/1 | 200.0 |
| **French Scores** | 0.0/100 | 69/100 | 38.677/100 | 0.942/1 | 33.33 |
| **Good Translation** | >50 | > 70 | > 50 | > 0.7 | < 30 |
| **Bad Translation** | < 15 | < 50 | < 30 | < 0.3 | > 70 |

**Table 4**. Example of how the translation metrics evaluate the Spanish and French back translations of "Tropical Weather Outlook", followed by reference score benchmarks identified by computational linguists as "good" or "bad" according to each metric.

All processed data, including translation scores and metadata, are visualized in VuGraf, an internally developed NWS/NESDIS Python-based data processing and visualization platform (Fig. 4). The dashboard provides real-time, interactive insights into translation quality and highlights areas needing attention. By drawing directly from AWS DynamoDB, it allows NWS staff to monitor and manage the translation of critical weather warnings with precision and efficiency. The dashboard remains under active development, with upcoming automation and other enhancements to be described in a future manuscript.

To supplement objective measures of translation quality, the public NWS translation website (see Section 5) also displays a "thumbs-up / thumbs-down" icon next to every translated product. Each response is logged in Google Analytics with the product's file name, message type, language, and timestamp. Translations receiving a negative rating are routed to human reviewers for correction, and the cumulative tallies quantify translation accuracy based on public perception of the translations across languages. The widget collects only this binary feedback, so it reflects sentiment without elaborating on specific translation issues. As these ratings are still being collected from the public via the translation website, we postpone statistical analysis of the rating dataset trends to future work as well.

## 3. Implementation of Translations Across NWS Areas of Responsibility

As the automated translation pilot with SJU concluded, the NWS expanded testing to evaluate the scalability and performance of the AI system across additional offices and languages. The NWS proceeded to conduct an analysis to choose WFOs where multilingual products would yield the greatest public benefit based on four ranking factors: (1) the number of distinct language groups in its County Warning Area (CWA), (2) the density of LEP residents, (3) the presence of local translation resources or partner organizations, and (4) the range of weather hazards the office routinely experiences. The third and fourth criteria came largely from internal NWS records; however, mapping the multilingual landscape proved more challenging and would require thorough analysis of the U.S. Census data.

To ensure our analysis contained reliable data, we used the U.S. Census Bureau's (2019) American Community Survey (ACS) 5-year estimates[4] from the C16001 Language Spoken at Home tables. Respondents to the ACS who reported speaking English 'less than very well'—including the categories 'Well,' 'Not Well,' and 'Not at All'—were used as a proxy for the LEP population in the United States. An initial evaluation focused on the linguistic needs of the entire nation to determine which languages to focus our efforts. To facilitate comparisons across languages, we calculated the ratio of the total number of speakers of each language (e.g., Spanish) to the subset with LEP (e.g., Spanish speakers who reported speaking English less than 'Very Well'). In our analysis, we prioritized languages with an LEP-to-total speaker ratio greater than 35% and an overall LEP population of at least 200,000. Based on these criteria, the NWS identified priority languages not simply by the size of their overall populations but by the likelihood that speakers may struggle to understand English-only warnings and thus require alerts in their dominant language. The list of priority languages as of 2026, presented alphabetically, includes Arabic, Chinese, French, Haitian Creole, Korean, Portuguese, Russian, Somali, Spanish, and Vietnamese. As new ACS data become available, this list will continue to evolve to reflect the nation's changing linguistic landscape.

To move from a national analysis to a more localized focus at select WFOs, we developed an interactive GIS dashboard that overlays U.S. Census data on multilingual populations with NWS

[4] The ACS is an annual nationwide survey from the U.S. Census Bureau that collects demographic, social, economic and housing characteristic data of the U.S. population. The 1-year ACS estimates are available for areas with populations greater than 65,000, whereas 5-years estimates are available to the block group level for all areas. Because of the higher accuracy and finer granularity, we opted for the 5-years estimates available in the Language Spoken at Home for Population Years and Over (C16001) table.

CWA boundaries to enhance operational applicability. Although ACS estimates are available at finer spatial scales, we adopted the CWA as our basic unit of analysis because the NWS issues warnings for the zones in the CWA rather than for individual counties. Most CWAs align with county boundaries, which allowed straightforward aggregation of county-level ACS data. However, in cases where a CWA bisected a county, we allocated that county's LEP population proportionally. For instance, if one CWA encompassed roughly one-third of a county and a neighboring CWA the remaining two-thirds, we first divided the county's LEP total in that ratio. Recognizing that LEP individuals are rarely distributed so evenly, we then replaced these coarse splits with census-tract counts wherever a county crossed multiple CWAs. Using tract-level data reduced spatial error and produced more accurate LEP estimates for the affected warning areas. A detailed discussion of this methodology and limitations of the dashboard can be found in Llewellyn et al. (2026).

After calculating LEP figures for each CWA, we uploaded the data to ArcGIS Online and created the Multilingual Community Visualizations (MCV) dashboard, an interactive tool that helps NWS personnel pinpoint the offices with the greatest translation needs (Fig. 5; Llewellyn et al. 2026). The dashboard combines LEP metrics and CWA boundaries, offering choropleth maps, filterable tables, and summary charts so practitioners and researchers can explore the information at different levels of detail.

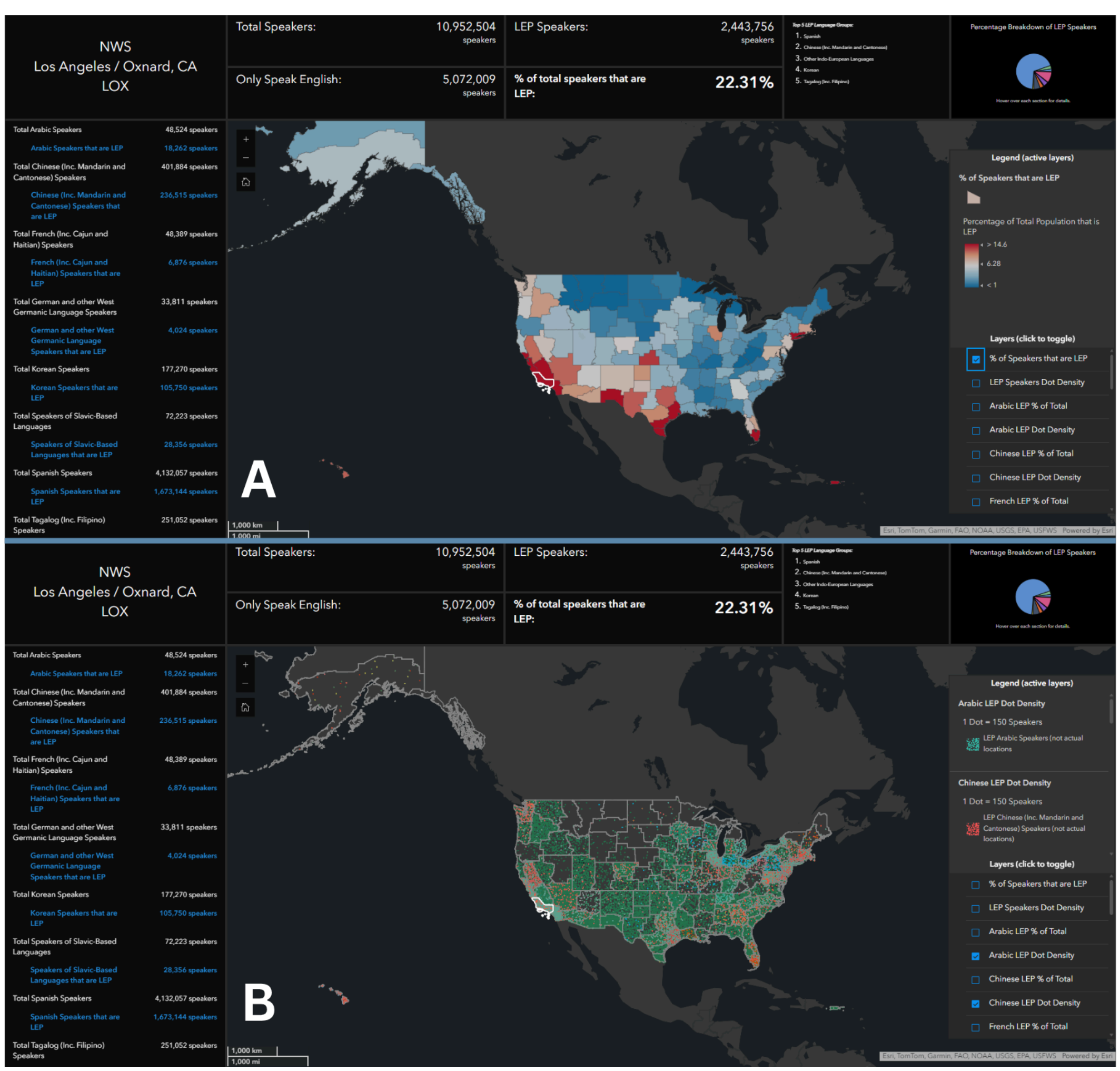

**Fig. 5.** The MCV dashboard displays national views of the A) percentage of the total CWA population that is LEP data layer and B) dot density of the LEP populations for multiple language layers within NWS CWA boundaries.

Upon accessing the dashboard, users are greeted with a national-level map view of linguistic information (Fig. 5), reflecting the dashboard's original design for internal NWS use (Llewellyn et al. 2026). When a specific CWA is selected, the dashboard dynamically updates to reflect relevant information for that area. The left panel of the dashboard displays the WFO identifier along with a general data viewer that provides updated language speaker counts for the selected CWA. At the top of the interface, an overview of key data is presented, complemented by a summary of the CWA's top 5 LEP language groups. Additionally, a pie chart illustrating the percentage breakdown of language groups is located in the top right corner, enhancing the visual representation of the data. To accommodate different analytical needs, the map window also offers both choropleth layers, which shade CWAs by LEP percentage, and dot-density layers, which depict the absolute size of each language group.

The criteria established, alongside the MCV dashboard, helped launch the AI Translation Program into its next experimental phase. As a result, the first public-facing translations were produced in the San Juan, Miami, Pago Pago, and New York City CWAs in 2023. Since then, multiple WFOs have participated in the NWS AI Translation Program (Fig. 6).

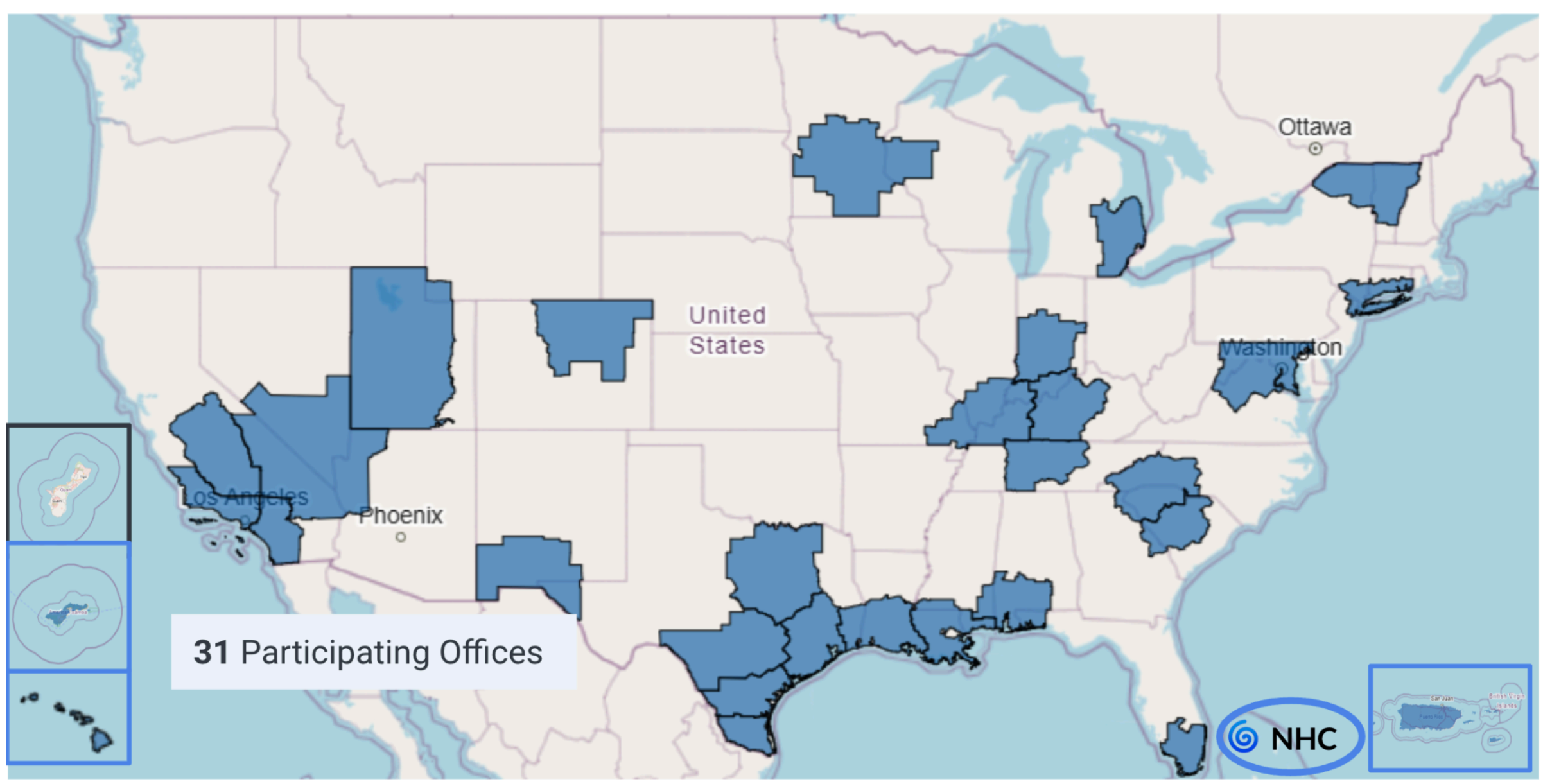

**Fig. 6**. NWS offices participating in the AI translation program are outlined in blue. 30 WFOs and the NHC are participating as of June 2025. Details on which languages an office provides translations can be found on the “About” page on the translation website.

## 4. Ethics and Trustworthiness in AI‑Based Weather Translation

Automated translations carry important societal implications, placing their development under heightened scrutiny. It is therefore crucial to define what we mean by ethics and trustworthiness in artificial intelligence, especially given that AI is not inherently ethically neutral but reflects the choices and values of its creators (Moorkens 2022). Ethical considerations center on the origins, quality, and appropriateness of training data, as well as transparency about how systems are built and deployed (Bostrom et al. 2024). Statistical models in machine translation inherently reflect the datasets on which they are trained, which means that issues of bias, copyright, consent, and privacy must be carefully managed to avoid perpetuating social inequalities (Canfora and Ottmann 2020; Lion et al. 2024; Moorkens 2022). Datasets must be selected with the model's intended application in mind, and developers in machine translation must clearly communicate domain limits, known biases, and intended uses to multilingual end users (Guerberof-Arenas and Moorkens 2023; Sakamoto 2019).

Trustworthiness, in turn, refers to qualities such as reliability, fairness, transparency, and accuracy (Bostrom et al. 2024). Trust is shaped by prior user experiences with the technology, the reputation of the AI developer, and developer transparency in communicating the processes used to create the technology (Chiou and Lee 2023). Repeated positive interactions with the technology can strengthen trust, while negative interactions can erode it quickly. In machine translation, model developers can bolster trust by involving users in the design process, providing channels for feedback, and maintaining human oversight through human-in-the-loop processes even after deployment (Federici et al. 2023; Guerberof-Arenas and Moorkens 2023). As Wirz et al. (2025) suggests, evaluating trustworthiness requires considering both the "trustor" (the user) and the "trustee" (the AI system), which means attention must extend beyond the system itself to the needs of its future users.

Applying these ethical principles to the NWS, this paper promotes transparency by detailing how the AI translation program was developed, trained, and evaluated to support trust among NWS forecasters and the broader weather, water, and climate enterprise. In this context, scientists are the primary users of the translation tool, while members of the public consume the translated information it produces. Our team of bilingual meteorologists, software developers, and translators deliberately addressed issues of ethics and trustworthiness in designing the automated translation tool, including the careful selection and preparation of training input data.

As discussed in section 2, we intentionally trained the model on archives of weather-related parallel texts and used the ongoing "balanced diet" methodology to capture phrases of unique messaging used during extreme weather events. Additionally, we engaged certified translators from the earliest design stages to select unbiased, dialect-neutral terminology accessible to all speakers, consistent with the recommendations of Trujillo-Falcón et al. (2021). These decisions ensured that the model's output within the constrained domain of weather vocabulary remained accurate and trustworthy (Lawson et al. 2025). All translations produced also underwent regular evaluation through back translation and other accuracy metrics (see Section 2c). This methodology allowed the NWS to quickly identify translation errors and promptly retrain the model as needed to maintain the reputation of the NWS as a trustworthy source of information, adhering to an ethical approach.

In our public-facing products, we clearly state that translations are generated using AI technology and include a disclaimer[5] noting that errors or omissions may occur. Omissions—when content from the source text is missing in the translated version—are especially important to acknowledge, as the translation may still read fluently and users might not realize that information is absent (Daems and Macken 2019). To foster public buy-in and collaboration with the program, we also invited feedback from the public themselves, providing multiple opportunities to submit feedback and have a say in how our translations could become the most effective, culturally representative and understandable. As mentioned in Section 2d, users can contribute a thumbs-up or thumbs-down rating for each translation or complete a short survey that allows for open-ended comments on our pilot website. These raw comments give us richer insight than binary ratings alone to guide ongoing model improvements and ideas for outreach content (Guerberof-Arenas and Moorkens 2023; Moorkens 2022).

While the public can be assured that the NWS AI Translation Program has been developed with themes of ethics and trustworthiness in mind, we acknowledge areas for improvement remain, and we anticipate future challenges as the program continues to evolve. As we expand the program into new languages, the NWS will gather extensive multilingual corpora in collaboration with native speakers and language experts to ensure consistent translation performance across all languages. This will be especially important for low-resource languages for which training data is not as easily accessible (Lion et al. 2024). Finally, we have begun social science research exploring public trust and perceptions of AI-generated translations across various contexts and scenarios (Gaviria Pabón et al. 2026). This social science research will help the organization better understand user's attitudes and needs to inform future adjustments or enhancements to NWS services.

## 5. A First Look At The Future, Multilingual National Weather Service

### *a. Introducing weather.gov/translate*

[5] The current disclaimer reads: "This product has been processed automatically using a translation program and may contain omissions and errors. The National Weather Service cannot guarantee the accuracy of the converted text. If there is any doubt, the English text is always the authorized version." This disclaimer is translated and displayed alongside multilingual products and has been reviewed and approved by NOAA's legal team.

The NWS AI Translation Program has made significant progress in the development and training of highly customized language models for weather contexts. However, the true life-saving potential of these curated models can only be fully realized when their translations are made accessible to the public. To bring our multilingual research and development closer to operations, the NWS released a new experimental translation website (weather.gov/translate) in October 2023, featuring translated text products and weather preparedness materials for our partners and the public (NOAA 2023). This site represents the first NWS platform to centrally host weather information in multiple non-English languages, developed and maintained through the NWS AI Language Translation Program, as the agency works toward full integration into its broader dissemination infrastructure.

The site was designed with LEP users in mind and is intentionally simple and mobile-friendly. Once users select their preferred language, the site displays all available content consistently in that language. Its main features include the following and are displayed in Fig. 7:

- **Home Page**: translated text products from participating NWS offices featuring blue pop-up glossary terms that provide context and indicate dialectal variations, along with a thumbs-up/thumbs-down rating system (Fig. 7A).
- **Forecast and Hazards Map**: an interactive map showing current NWS translated watches, warnings and advisories, with the option to zoom to local conditions and view a 7-day point-and-click forecast. Additional map layers include radar, fronts, and SPC/WPC/NHC outlooks, NHC tropical cyclone tracks and cone of uncertainty in addition to HeatRisk index (Fig. 7C).
- **Product Descriptions**: plain-language explanations of NWS text products, their severity, and when they are typically issued.
- **Infographics**: translated graphics that convey preparedness information in a visual format, often shared by WFOs and partners on social media (Fig. 7B). These graphics are produced in both English and other languages, enabling partners to post parallel content and reach wider audiences.
- **About**: background on the AI Translation Program process and a map of participating offices.
- **Feedback**: surveys available in multiple languages that allow the public to share comments directly with the NWS.

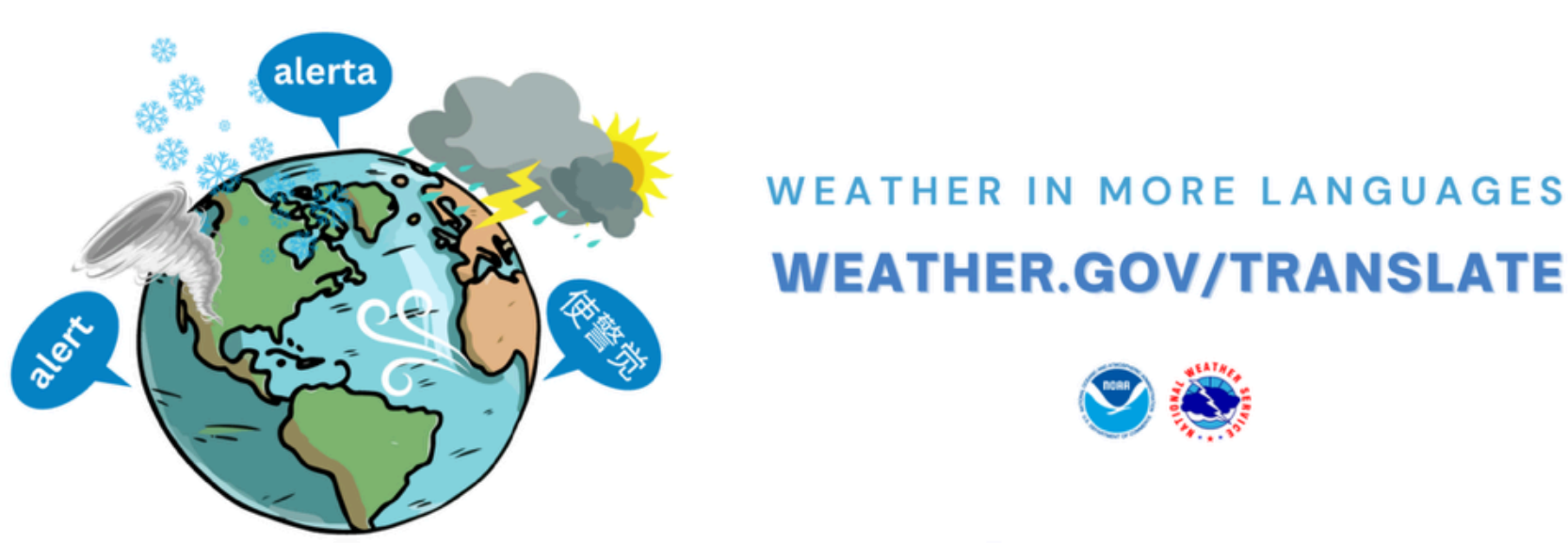

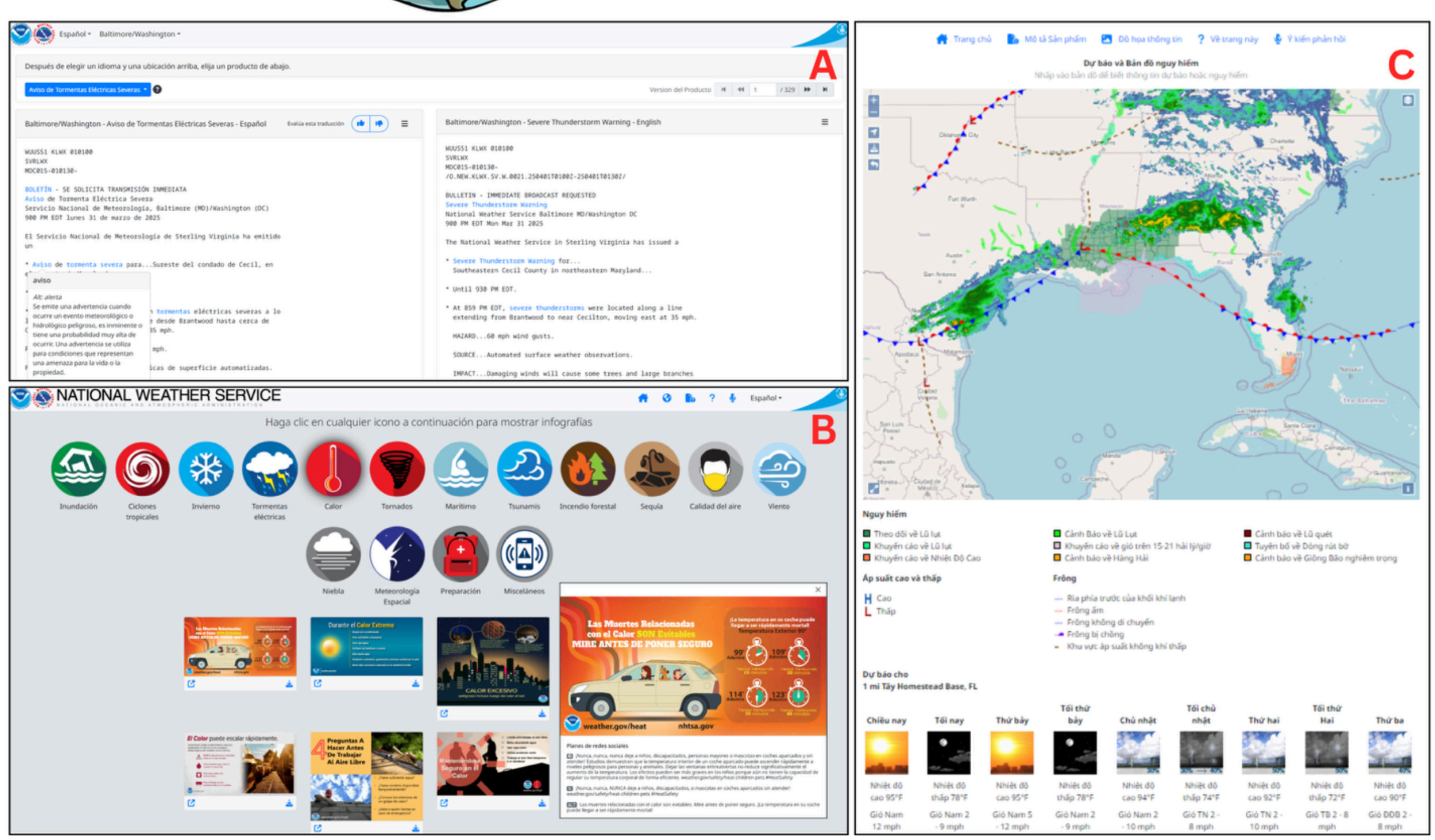

**Fig. 7.** Main visual pages of weather.gov/translate. (A) Homepage with a Spanish Severe Thunderstorm Warning from the Baltimore/Washington office, highlighting the pop-up glossary feature. (B) Infographics page organized by hazard icons, with expanded graphics showing translated captions and alternate text. (C) Forecast and Hazards Map with current NWS watches and warnings, a 7-day forecast in Vietnamese, and overlays of radar and weather fronts. A PDF version of this figure is available at https://tinyurl.com/nwstranslate.

Many WFOs have adopted the AI-assisted translation capabilities to better translate full-text NWS products for their local LEP communities, reducing the need for manual content creation and translation. Multilingual infographics have moved from experimentation to operational use, now appearing in NWS graphics, national campaigns like Hurricane Preparedness Week, and across NWS social media during high-impact events (e.g., Fig. 8, Fig. A1, and Fig. A2). These advances illustrate how multilingual products are becoming embedded in NWS operations, supporting both local, regional, and national messaging during extreme weather events. Building upon this momentum, after testing with the AI solution, the NHC announced that beginning with the 2025 hurricane season, it will officially provide AI-translated Spanish versions of the Tropical Weather Outlook, Public Advisories, Tropical Cyclone Discussion, Tropical Cyclone Update, and Key Messages (NOAA 2025).

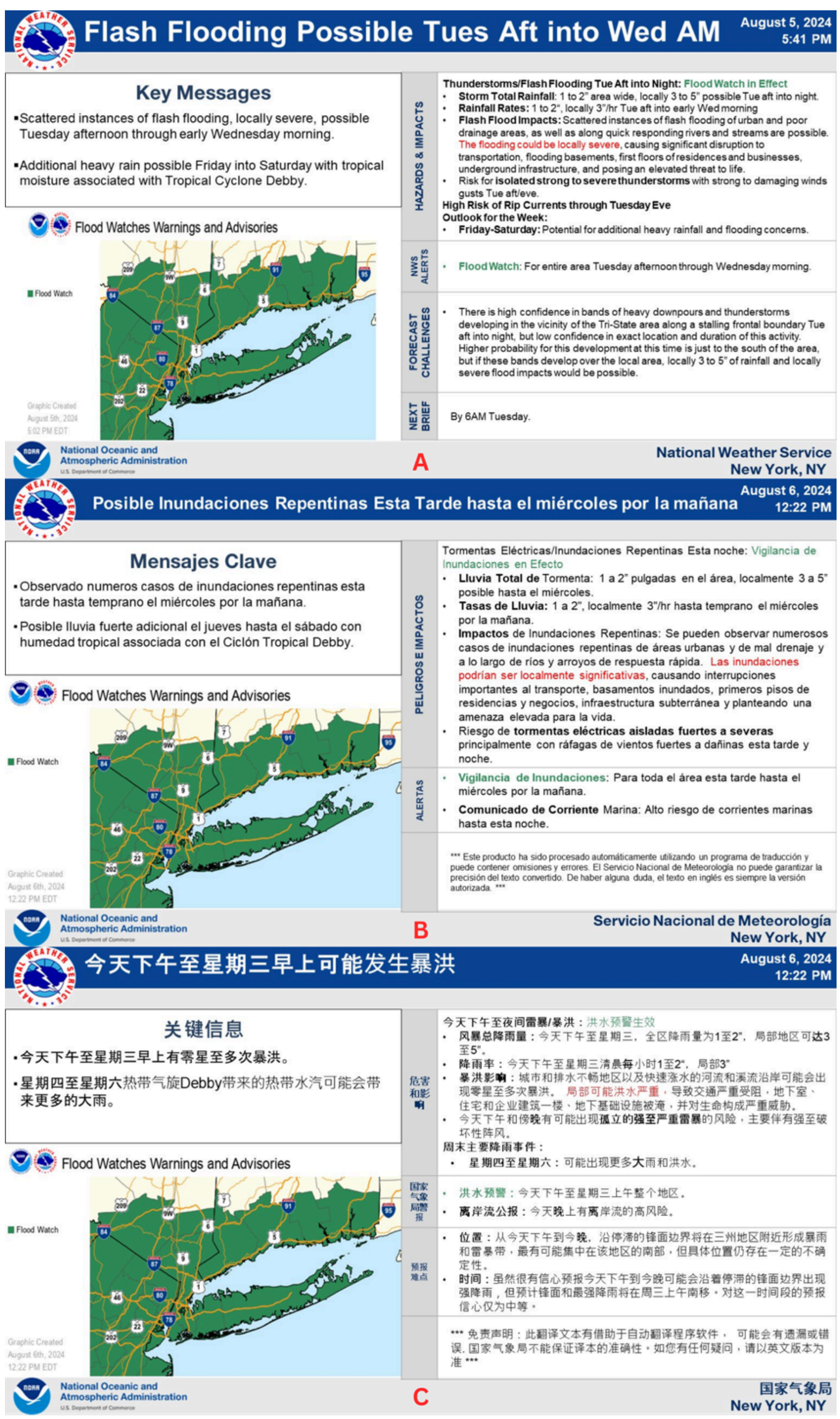

**Fig. 8.** Example of a multilingual weather briefing shared by NWS New York, NY WFO on social media during a flooding event. The briefing is shown in English (A), Spanish (B), and Chinese (C) using translations generated through the NWS AI Language Translation Program. A PDF version of this figure is available at https://tinyurl.com/nycfloodtranslation.

An interesting point of note, however, is that during Hurricane Beryl (2024), a fully Spanish post was mistranslated into English by Facebook's auto-translate feature (Fig. 9; https://www.facebook.com/share/p/17MgHDP6Rz/), while the NHC was testing its AI translation system. The authors speculate that Facebook's auto-translator was likely not trained on proper terms for weather phenomena across multiple languages and therefore mistranslated the correct Spanish term for "storm surge" (i.e., "marejada ciclónica") several times within the post. These mistranslations confused only the English-speaking Facebook users viewing the Spanish post in their feeds—especially the egregious "cyclonic dizziness" error—which drew both alarmed and humorous comments. This cautionary tale shows that even when the original message is accurate, downstream, untrained automated tools can distort meaning and create public confusion, highlighting the need for trusted sources like the NWS and NHC to utilize and invest in customized automated translation services for effective, life-saving communication.

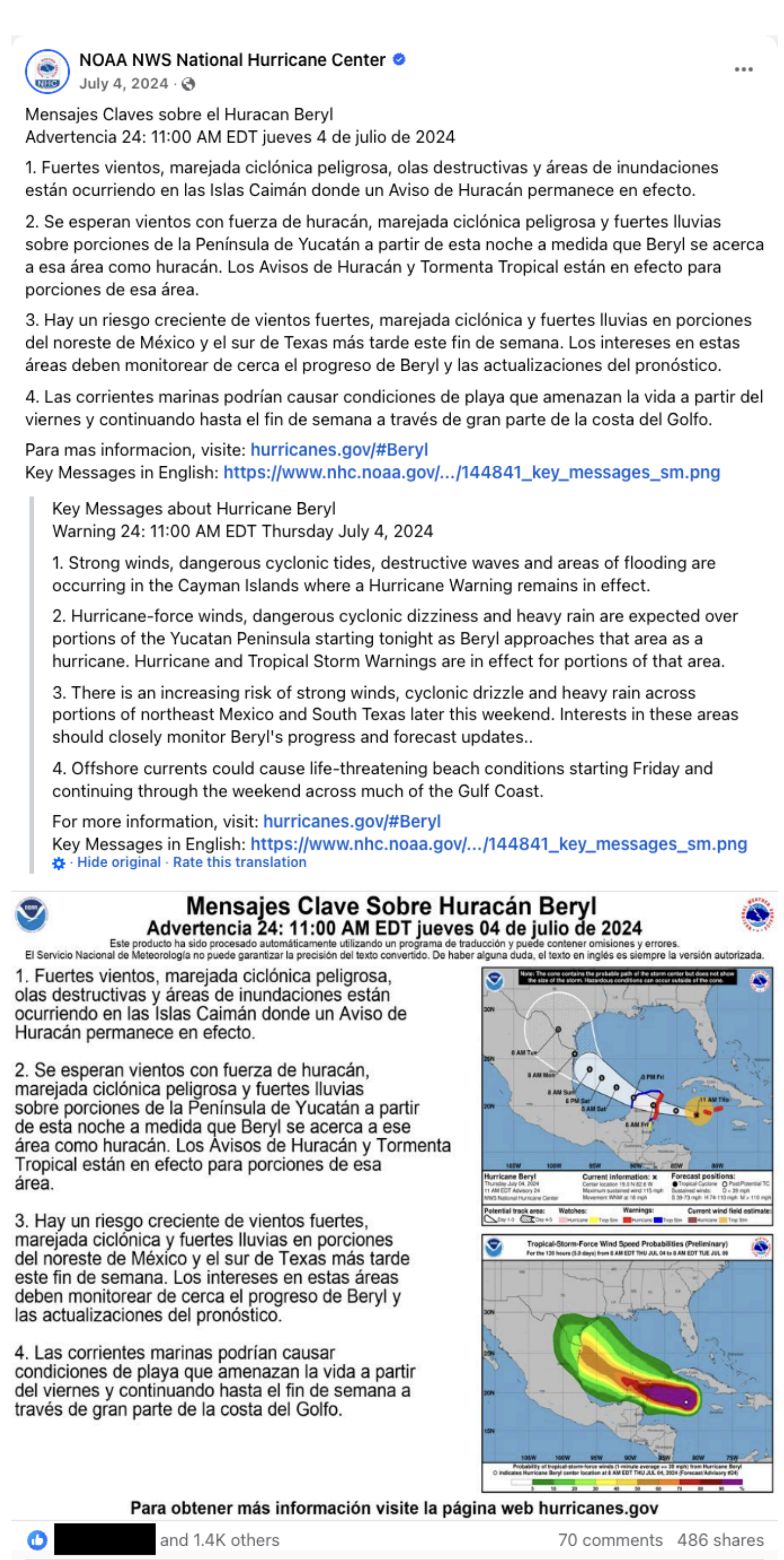

**Fig. 9.** An NHC Spanish Key Messages Facebook post about Hurricane Beryl where Facebook's auto-translation feature mis-translated the correct Spanish term for "storm surge" causing confusion by the English-speaking audience.

*b. Future Directions*

Looking ahead, the NWS has outlined multiple milestones to strengthen and expand its AI Language Translation Program. In the near term, the agency aims to run the entire translation process from the cloud, allowing the NWS to rapidly expand translation sites onto the LILT system. Efforts are also underway to increase the number of translated infographics available on the translation website, integrate multiple additional languages into the Hazards Map page, and expand internal access to the NWS-trained language models that can assist with official NWS documents and weather briefing translations. These initial steps are intended to scale current successes while making sure that translated products are available, accessible, and visible across NWS platforms.

Beyond these immediate measures, the NWS has identified outcome-oriented goals that will shape the long-term trajectory of multilingual communication. Current priorities include translated partner weather briefings for major events such as the World Cup, along with new back translation techniques to validate accuracy and improve LLM performance. The NWS also aims to centralize terminology through an online multilingual dictionary, while developing a fully translated weather.gov and integrating translations into the NWS API. These API upgrades will make it easier for operational partners, including media outlets, emergency management agencies, and private sector weather organizations, to access and disseminate translated NWS content within their own platforms. As this work continues, collaboration with Spanish-language broadcast meteorologists and app developers could provide valuable insight into audience-centered language practices that further support comprehension and weather literacy. The NWS AI Translation Program is also exploring ways to address literacy and accessibility concerns, including adding audio recordings to accompany translated text warnings and accessing whether AI can deliver rapid weather warnings in American Sign Language (ASL). Looking ahead, the roadmap may also incorporate intentionally tailored multilingual outreach, such as comprehensive campaigns shared through dedicated NWS multilingual social media accounts in addition to funded national research surveys that invite public participation for providing in-depth feedback.

Achieving these goals will likely depend on continued investments in funding, workforce capacity, and IT infrastructure. Potential areas of support could include dedicated staff or contractors to help oversee the AI Translation Program, as well as mechanisms that enable bilingual NWS staff to contribute their expertise in reviewing and validating translations. Partnerships with cultural heritage groups and academic institutions, particularly multilingual students and faculty, could also provide valuable opportunities to collaborate with NWS staff on analyzing and refining AI translations.

At the same time, the agency faces structural limitations in dissemination technology that must be resolved before multilingual information can be transmitted nationwide. Currently, NWS systems rely on ASCII text encoding, which only accommodates the most common Latin-based characters. This creates challenges for Asian languages and even for some accented characters in European languages, where intended messages may be simplified to avoid characters that cannot be transmitted, as is currently the case for the Spanish Wireless Emergency Alerts (WEAs) transmitted by the NWS (Trujillo-Falcón 2024). To address this, the agency has prioritized accuracy in translation during LLM training while also working toward longer-term solutions that incorporate the UTF-8 standard, which enables computers to process text from any language. Another closely related challenge is the need to package English and multilingual products together with trackable headers so that both can be disseminated simultaneously, a technical hurdle that remains unresolved within both the NWS and the WMO.

Finally, the authors personally acknowledge several internal and external challenges that could influence progress toward these milestones. Current funding structures make it difficult for university partners to secure sustained support for translation research, which in turn limits opportunities for NOAA collaboration. Travel restrictions further constrain outreach and cooperation with local and international partners. At the systems level, FEMA's Integrated Public and Alert Warning System, or IPAWS, will require updates to accommodate multilingual characters and/or UTF-8 compliance across both the WEA and Emergency Alert System. Additionally, despite the significant progress already achieved by the NWS, it is not yet possible to fully meet the translation needs of every at-risk LEP community. While LLMs are being customized for major global languages, indigenous and Alaskan native languages present unique challenges given their primarily oral nature and the absence of sufficiently large digital or written corpora necessary for model development. To ensure these cultures do not face being uninformed and under prepared when hazardous weather strikes, this will likely require collaboration with AI language service providers as well as significant investment and partnership with cultural groups to preserve these languages and make them accessible for technological applications. Overcoming these hurdles, alongside modernizing dissemination technology, will be essential for moving from experimental pilots to a fully operational nationwide system for multilingual weather communication.

## 6. Conclusion

The NWS AI Language Translation Program marks a turning point in the effort to make lifesaving weather information accessible to all. By integrating AI with human expertise, the agency has begun to move from ad hoc translations toward a scalable, systematic approach that can serve diverse language communities nationwide. Challenges remain, especially in dissemination technology and long-term support, but the progress to date demonstrates that multilingual communication is both feasible and necessary. Weather warnings are most effective when people can understand them, and this program lays the groundwork for a truly effective *Weather-Ready Nation, Nación preparada para el tiempo,* or *天气准备就绪的国家*.

A*cknowledgments.*

We would like to acknowledge the various translation teams across the NWS that pioneered and advanced multilingual communication to where it is today. These include NWS SJU, Multimedia Assistance in Spanish (MAS), and the Spanish Outreach Team (SOT). This material is based on research supported by the Cooperative Institute for Severe and High-Impact Weather Research and Operations Director's Discretionary Research Fund and the Joint Technology Transfer Initiative Program within the NOAA/OAR Weather Program Office under Award #NA22OAR4590187, who also funded J.E.T.F. and S.T.H. The opinions and conclusions contained in this paper are those of the authors and should not be interpreted as representing the official policies, either expressed or implied, of NOAA, the NWS or the U.S. Department of Commerce.

*Data Availability Statement.*

Open-source Python packages used for the linguistic metrics in Table 2 are available at the following links: SacreBLEU for BLEU scores (https://github.com/mjpost/sacrebleu), TER and ChrF++, TheFuzz for fuzzy string matching (https://github.com/seatgeek/thefuzz), and COMET (https://github.com/Unbabel/COMET). The MCV dashboard is available in Llewellyn et al. (2026) or at https://experience.arcgis.com/experience/71664444b7a74403abc592967375c3a9. The experimental public website can be accessed at https://weather.gov/translate.

APPENDIX A

*Examples of Multilingual Social Media Communications from NWS WFOs*

Figs. A1 and A2 provide additional examples of social media posts and public-facing communications disseminated by NWS WFOs using translations produced through the NWS AI Language Translation Program.

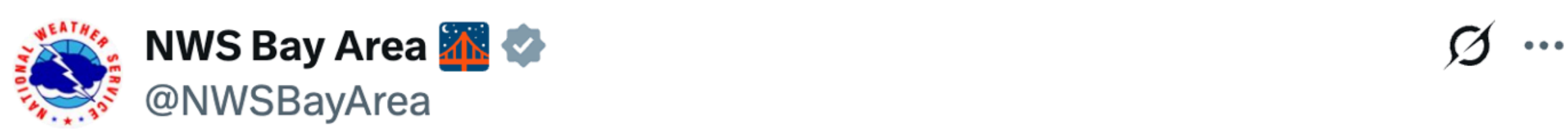

🌡️Dangerous and potentially lethal heat continues today! If going outdoors, drink plenty of water, wear light-colored clothing, and take frequent breaks in the shade! weather.gov/safety/heat #CAwx

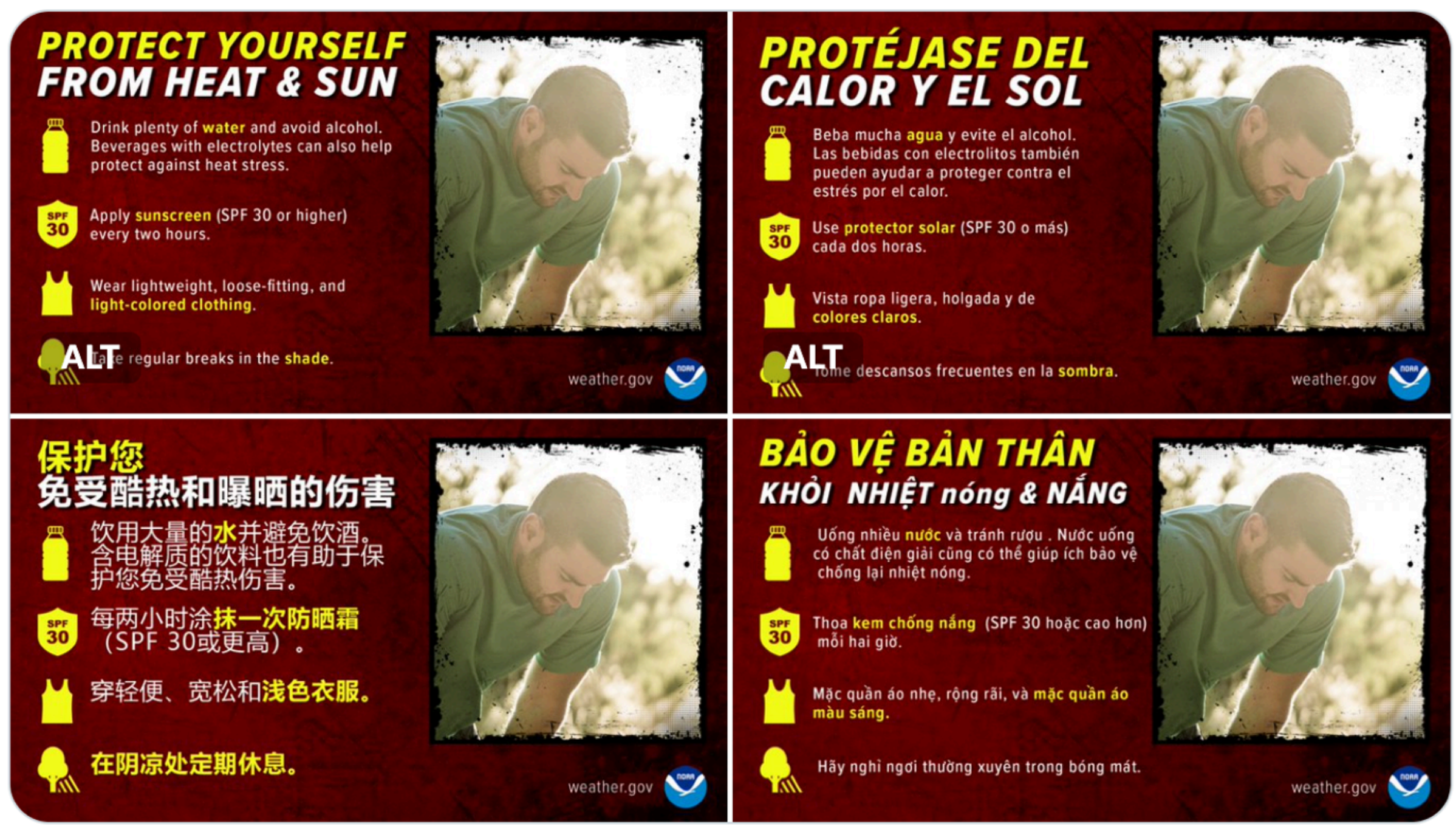

7:20 AM · Jul 3, 2024 · **10K** Views

**Fig. A1.** Example of a multilingual heat safety graphic shared by the NWS San Francisco Bay Area, CA WFO on social media during an extreme heat event (https://x.com/NWSBayArea/status/1808475844039594026?s=20). The post includes the same heat safety guidance presented in English, Spanish, Chinese, and Vietnamese using translations produced through the NWS AI Language Translation Program.

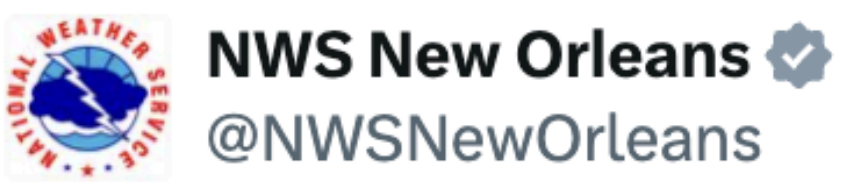

Show translation

DYK our warnings are online in Spanish and Vietnamese?

¿Sabías que nuestras advertencias están disponibles en línea en español? Consulta los gráficos a continuación.

Bạn có biết cảnh báo của chúng tôi có sẵn trực tuyến bằng tiếng Việt không? Xem hình ảnh bên dưới.

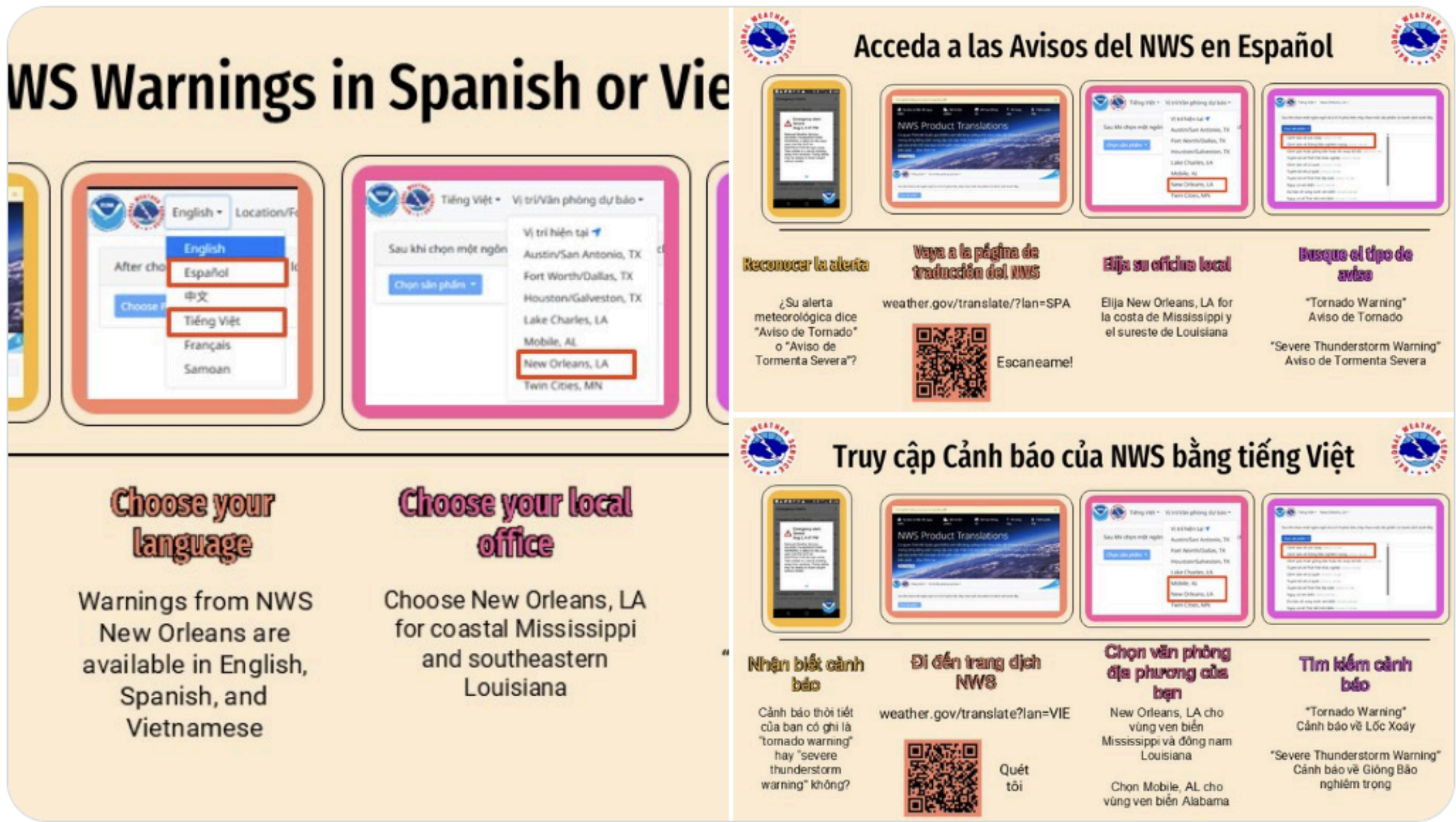

3:41 PM · Mar 14, 2025 · **3,100** Views

**Fig. A2.** Multilingual outreach post from the NWS New Orleans, LA WFO informing the public about the availability of official NWS warning products in English, Spanish, and Vietnamese (https://x.com/NWSNewOrleans/status/1900648414410318286). The post guides readers to translate warnings provided through weather.gov/translate